\documentclass{article}
\usepackage{spconf,amsmath,epsfig}

\usepackage{graphicx}
\usepackage{url}
\usepackage{amsmath}
\usepackage{amsfonts}
\usepackage{xcolor}
\usepackage{subdepth} 

\usepackage{cite}

\newcommand{\x}{\mathbf{x}}
\newcommand{\z}{\mathbf{z}}
\newcommand{\w}{\mathbf{w}}
\renewcommand{\t}{\mathbf{t}}
\def\T{\mathsf{T}}

\newcommand{\bmu}{\mbox{\boldmath $\mu$}}

\newcommand{\eq}[1]{Eq.~(\ref{eq:#1})}

\newcommand{\fig}[1]{Fig.~\ref{fig:#1}}

\usepackage{xcolor}
\newcommand{\remark}[1]{\textcolor{red}{[Remark: #1]}}
\renewcommand{\remark}[1]{}

\title{CLOSED-FORM DETECTOR FOR SOLID SUB-PIXEL TARGETS\\ IN 
  MULTIVARIATE $t$-DISTRIBUTED BACKGROUND CLUTTER}

\name{James Theiler$^a$, Beate Zimmer$^b$, and Amanda Ziemann$^a$\thanks{This work was supported in part by the United States Department of Energy NA-22 project on Hyperspectral Advanced Research and Development for Solid Materials (HARD Solids).}}
\address{$^a$Space Data Science and Systems Group, Los Alamos National Laboratory,
Los Alamos, NM\\
$^b$Department of Mathematics and Statistics, Texas A\&M University--Corpus Christi, Corpus Christi, TX}
\begin{document}
%
\maketitle
\begin{abstract}
The generalized likelihood ratio test (GLRT) is used to derive a detector for solid sub-pixel targets in hyperspectral imagery.  A closed-form solution is obtained that optimizes the replacement target model when the background is a fat-tailed elliptically-contoured multivariate $t$-distribution. This generalizes GLRT-based detectors that have previously been derived for the replacement target model with Gaussian background, and for the additive target model with an elliptically-contoured background.  Experiments with simulated hyperspectral data illustrate the performance of this detector in various parameter regimes.
\end{abstract}
\section{Introduction}
\label{sec:intro}

To detect small targets in cluttered backgrounds requires models of
the target, of the background, and of how the two interact.  Although
target variability models are important, particularly for solid
targets\cite{Myers15}, 
we will take the target signature as a single vector.  However, we
will consider a range of elliptically-contoured background models,
from Gaussian to very fat tailed, and we will consider two different
target-background interaction models -- the additive model and the
replacement model -- that incorporate variability into the
\emph{strength} of the target.

\subsection{Background models}

The importance of background modeling has recently been
emphasized\cite{Matteoli14}, and although the Gaussian model is often
surprisingly effective, a useful extension is the multivariate
$t$-distribution, which has in particular been proposed for
hyperspectral imagery\cite{Manolakis01}.  This is similar to the
Gaussian in that it is defined by a mean and a covariance matrix,
which implies that the distribution is uni-modal with ellipsoidal
contours of constant density; this density decreases with distance from
the mean, but the decrease can be much slower than the $\exp(-r^2)$
decay exhibited by Gaussians, leading to heavier tails that are often
more representative of observed data.

\subsection{Target-background interaction models}

In many signal detection applications, the signal of interest is
assumed to be additive with respect to a background that is generally
characterized in some statistical way.  We can write
\begin{equation}
  \label{eq:add}
  \x = \z + \alpha\t
\end{equation}
where $\x\in\mathbb{R}^d$ is the measured signal, $\z\in\mathbb{R}^d$ is the
background signal, $\t\in\mathbb{R}^d$ is the signal of interest, $d$ is the number
of spectral channels, and $\alpha$ is a
scalar quantity that characterizes the strength of the signal. The
additive model is the basis for many traditional target detection
algorithms, including the adaptive matched filter (AMF)\cite{Reed74} and
the adaptive coherence estimator (ACE)\cite{Kraut05}.  When the
background is multivariate $t$, then the GLRT solution for the additive
model leads to a detector we will here call the elliptically-contoured
adaptive matched filter\footnote{In \cite{Theiler08igarss}, it is given a different
name (EC-GLRT) that is not as consistent with the naming conventions used
in this paper.} (EC-AMF)~\cite{Theiler08igarss}; it is given by
\begin{equation}
  \label{eq:ecmf}
  \mathcal{D}(\x) = \frac{\sqrt{(\nu-1)}\,\,\t^\T R^{-1}(\x-\bmu)}{\sqrt{(\nu-2)+(\x-\bmu)^\T R^{-1}(\x-\bmu)}},
\end{equation}
where $\bmu$ is the mean and $R$ is the covariance matrix of the background distribution.
In \eq{ecmf}, $\nu\to\infty$ leads to the AMF detector, and $\nu\to 2$ leads to the ACE detector.

While the additive model is the basis for many target detection algorithms,
it has limitations\cite{Schaum14}, and in particular does not account
for the occlusion of the background by a solid target.  In the
replacement model\cite{Stocker97,Manolakis01-tgrs}, we treat $0\le\alpha\le
1$ as the target area (fraction of a pixel), and write
\begin{equation}
  \label{eq:replace}
  \x = (1-\alpha)\z + \alpha\t.
\end{equation}
This is called the replacement model because a fraction $\alpha$ of
the background signal~$\z$ is \emph{replaced} with target signal~$\t$.
The finite target matched filter (FTMF), derived by Schaum and
Stocker\cite{Schaum97}, is the replacement-model version of the AMF:
it is the GLRT solution to \eq{replace} in the
situation that the background~$\z$ is Gaussian.  Although this
detector is somewhat more complicated than the AMF, or even the EC-AMF
in \eq{ecmf}, it can be written as a closed-form expression.

Closed-form generalizations of the FTMF have been derived for Gaussian
target variability\cite{DiPietro10} and alternative models of
covariance scaling\cite{Schaum14ao}. Here, we derive a
closed-form GLRT solution when the background is a general class of
elliptically-contoured distribution.  In the special case that the
background is Gaussian, we obtain the FTMF solution.

We treat the target detection problem in a hypothesis testing
framework, with the null hypothesis corresponding to $\alpha=0$ and
the alternative associated with $\alpha>0$.  Since the nonzero
$\alpha$ is unspecified, this is a composite hypothesis testing
problem\cite{Lehmann05}, and we use the 
generalized likelihood ratio test (GLRT) to derive our detector.  The
detector is a function of $\x$ given by the logarithm of this ratio
\begin{equation}
  \label{eq:glrt}
  \mathcal{D}(\x) = \log\,\frac{\max_\alpha p_x(\x|\alpha)}{p_x(\x|0)}.
\end{equation}
The expression in \eq{glrt} is written in terms of $p_x$, which is 
the probability density function for $\x$. We can express this function in 
terms of $p_z(\z)$, the probability
density associated with the background $\z$. We have
\begin{equation}
  \label{eq:glrtz}
  p_x(\x|\alpha) = (1-\alpha)^{-d}p_z((\x-\alpha\t)/(1-\alpha))
\end{equation}
where the argument $(\x-\alpha\t)/(1-\alpha)$ is obtained by solving \eq{replace}
for $\z$, and 
where the prefactor $(1-\alpha)^{-d}$ arises from the Jacobian of the transformation
of variables from $p_x$ to $p_z$.


Taking $\z$ to have mean $\bmu$ and covariance $R$, the multivariate $t$
distribution is given by
\begin{equation}
  \label{eq:mt}
  p_z(\z) = c\,|R|^{-d/2}\left(1+\frac{(\z-\bmu)^\T R^{-1}(\z-\bmu)}{\nu-2}\right)^{-\frac{d+\nu}{2}}
\end{equation}
where $d$ is number of spectral channels, $\nu$ is a parameter that
specifies how fat-tailed the distribution is (larger $\nu$ is less
fat-tailed, with the $\nu\to\infty$ limit corresponding to a Gaussian
distribution), and the normalizing constant $c$ depends only on $d$ and~$\nu$.
Thus, 
\begin{align}
  p_x(\x|\alpha) =& (1-\alpha)^{-d}p_z((\x-\alpha\t)/(1-\alpha)) \\
                 =& \frac{c\,|R|^{-d/2}}{(1-\alpha)^{d}}
               \left(1+\frac{\w^\T R^{-1}\w}{(1-\alpha)^2(\nu-2)}\right)^{-\frac{d+\nu}{2}}
  \label{eq:pzxa}
\end{align}
where $\w = (\x-\mu) - \alpha(\t-\mu)$.
To find the value of $\alpha$ that maximizes \eq{pzxa}, we can 
take the derivative of $\log p(\x|\alpha)$ with respect to
$\alpha$, set that expression to zero, and solve for~$\alpha$.
For Gaussian $p_z(\z)$, that approach was found\cite{Schaum97}
to produce a quadratic equation in $\alpha$.  For the more general multivariate $t$-distribution, 
we also obtain a quadratic equation, though with modified coefficients.
The solution to that quadratic equation is given by 
\begin{equation}
  \label{eq:ml-alpha}
  \widehat\alpha = 1 - \frac{-B+\sqrt{B^2-4AC}}{2A}
\end{equation}
where
\begin{align}
  A &= (\t-\bmu)^\T R^{-1}(\t-\bmu) + (\nu-2), \label{eq:ml-alpha-A}\\
  B &= (1-\nu/d)(\x-\t)^\T R^{-1} (\t-\bmu), \\
  C &= -(\nu/d)(\x-\t)^\T R^{-1} (\x-\t). \label{eq:ml-alpha-C}
\end{align}
This value of $\alpha$ satisfies $p_x(\x|\widehat\alpha) = \max_\alpha p_x(\x|\alpha)$.
Thus, our detector, the elliptically-contoured finite target matched filter (EC-FTMF), is given by
\begin{equation}
  \label{eq:ec-ftmf-detector}
  \mathcal{D}(\x) = \log p_x(\x|\widehat\alpha) - \log p_x(\x|0)
\end{equation}
with $p_x(\x|\alpha)$ given in \eq{pzxa} and $\widehat\alpha$ given by Eqs.~(\ref{eq:ml-alpha}-\ref{eq:ml-alpha-C}).

\begin{table}[t]
  \caption{\label{tab:detectors}Taxonomy of detection algorithms. The EC-FTMF (and its special
    case FTCE) are introduced in this paper, to extend the FTMF algorithm to 
    non-Gaussian backgrounds.}
\begin{center}
\begin{tabular}{|r|ccc|}
\hline
& Gaussian & Multivariate $t$ & Fat-tailed \\
Target model & $\nu\to\infty$ & $2\le\nu\le\infty$ & $\nu\to 2$ \\
\hline
Additive & AMF\cite{Reed74} & EC-AMF\cite{Theiler08igarss} & ACE\cite{Kraut05} \\
Replacement & FTMF\cite{Schaum97} & EC-FTMF & FTCE \\
\hline
\end{tabular}
\end{center}
\end{table}

In the $\nu\to\infty$ limit, the multivariate t becomes Gaussian, and
the expressions in Eqs.~(\ref{eq:ml-alpha-A}-\ref{eq:ml-alpha-C})
diverge.  But in \eq{ml-alpha} it is only the relative values that
matter; thus we can express this limit with the expressions
\begin{align}
  B/A &= -(\x-\t)^\T R^{-1} (\t-\bmu)/d, \\
  C/A &= -(\x-\t)^\T R^{-1} (\x-\t)/d.
\end{align}
These values recapitulate the FTMF result obtained for a Gaussian background\cite{Schaum97}.
\remark{In particular, $B/A$ here is $\beta-\gamma$ in \cite{Schaum97}; similarly, $C/A$ here
is $\alpha-2\beta+\gamma$ in \cite{Schaum97}.}

For $\nu\to 2$, we have the heavy-tailed limit
\begin{align}
  A &= (\t-\bmu)^\T R^{-1}(\t-\bmu), \\
  B &= (1-2/d)(\x-\t)^\T R^{-1} (\t-\bmu), \\
  C &= -(2/d)(\x-\t)^\T R^{-1} (\x-\t),
\end{align}
which we call the finite target coherence estimator (FTCE).

We remark that these three replacement-model detectors, the general EC-FTMF and the
special cases FTMF and FTCE, have corresponding detectors associated
with the additive model in \eq{add}, as shown in Table~\ref{tab:detectors}.  These
additive-model detectors are the EC-AMF
and its special cases, AMF and ACE.
For very small $\alpha$ and very
large target magnitude $|\t|$, we expect these replacement-model
detectors to be well approximated by their associated
additive-model detectors.  In this sense, we can argue that the
EC-FTMF detector described in \eq{ec-ftmf-detector} covers all six
cases.

\section{Simulation}

\begin{figure}[t]
\centerline{\includegraphics[width=0.9\columnwidth]{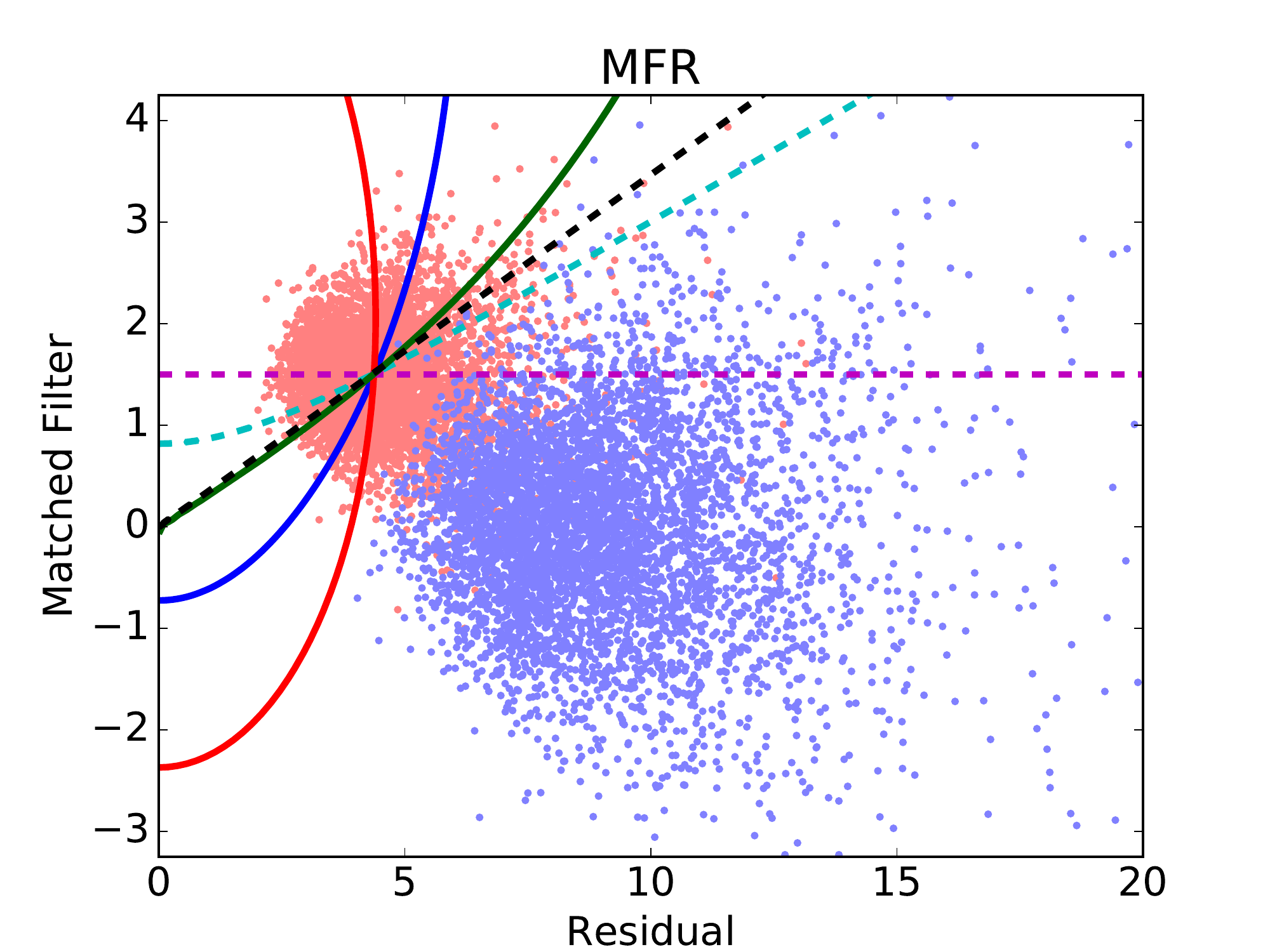}}
\centerline{\includegraphics[width=0.9\columnwidth]{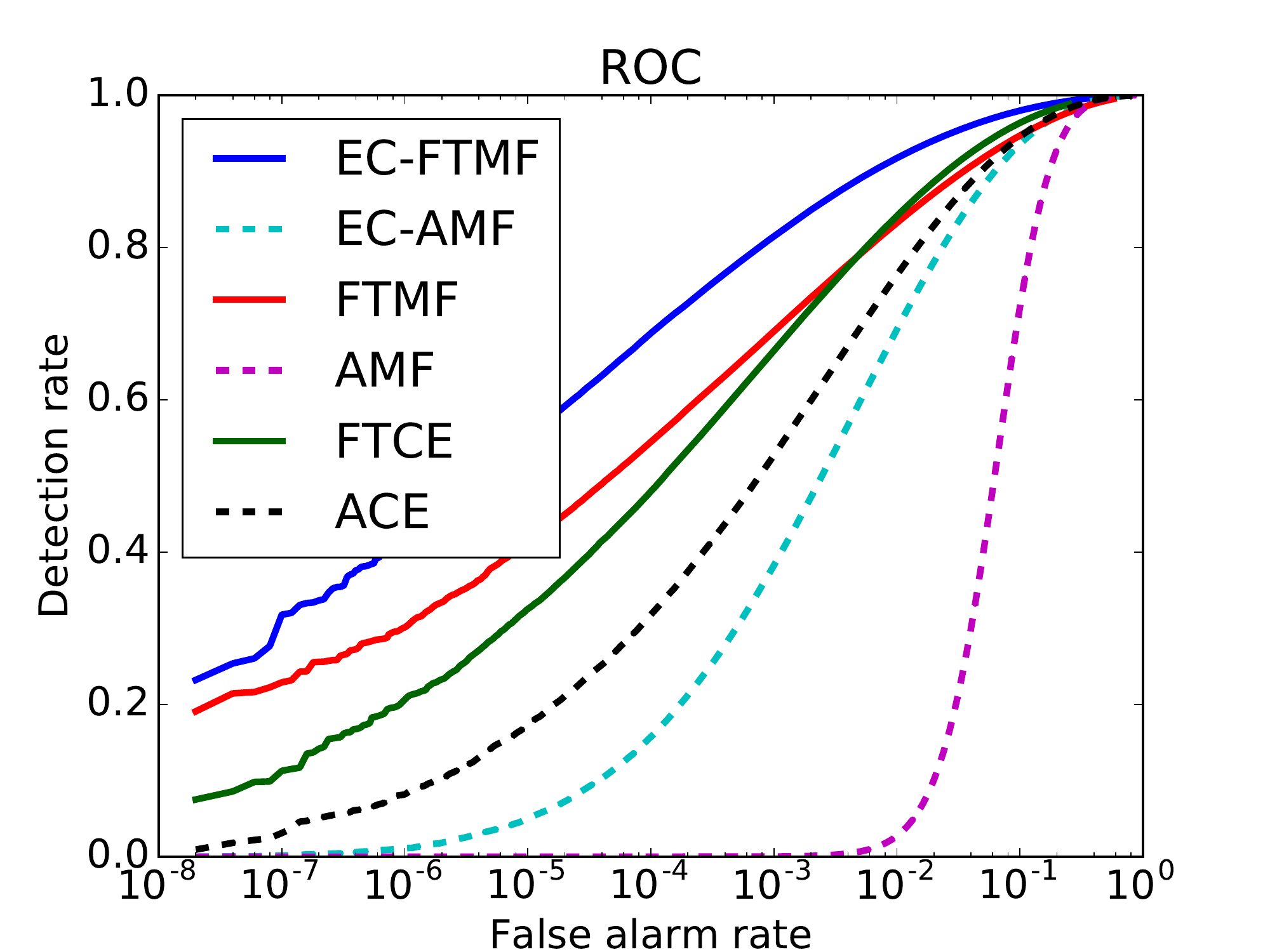}}
\caption{\label{fig:mfrroc}Top panel is Matched-Filter Residual (MFR) plot of
  simulated data, showing both background (blue) and target (red)
  pixels, along with the contours associated with several different
  detection algorithms.  The contours are chosen so that the detection
  rate is exactly 0.5; the better detectors are those with fewer false
  alarms, which are associated with blue pixels that are ``above'' the
  contours. Bottom panel is Receiver Operating Characteristic (ROC) curves for 
  these detectors.  Here, $\nu=10$, $d=90$, $T=3$, and $\alpha=0.5$.}
\end{figure}

We can illustrate the performance of the EC-FTMF detector on simulated data.
In this simulation we draw $N$ samples from a $d$-dimensional
multivariate $t$-distribution parameterized by $\nu$, with (for
simplicity) zero mean and unit covariance.  These $N$ samples are
representative of background pixels from a multi- or hyper-spectral
image that have been de-meaned and whitened.

For each background sample, we used the matched-pair
formulation\cite{Theiler13mpml,Theiler14spie} to produce
an associated target pixel, produced by the replacement model in
\eq{replace} using a fixed value of $\alpha$ (which we know, but the
algorithm does not).  Our target signature $\t$ is given by a vector
of magnitude $T$.

\begin{figure}[t]
\centerline{\includegraphics[width=0.9\columnwidth]{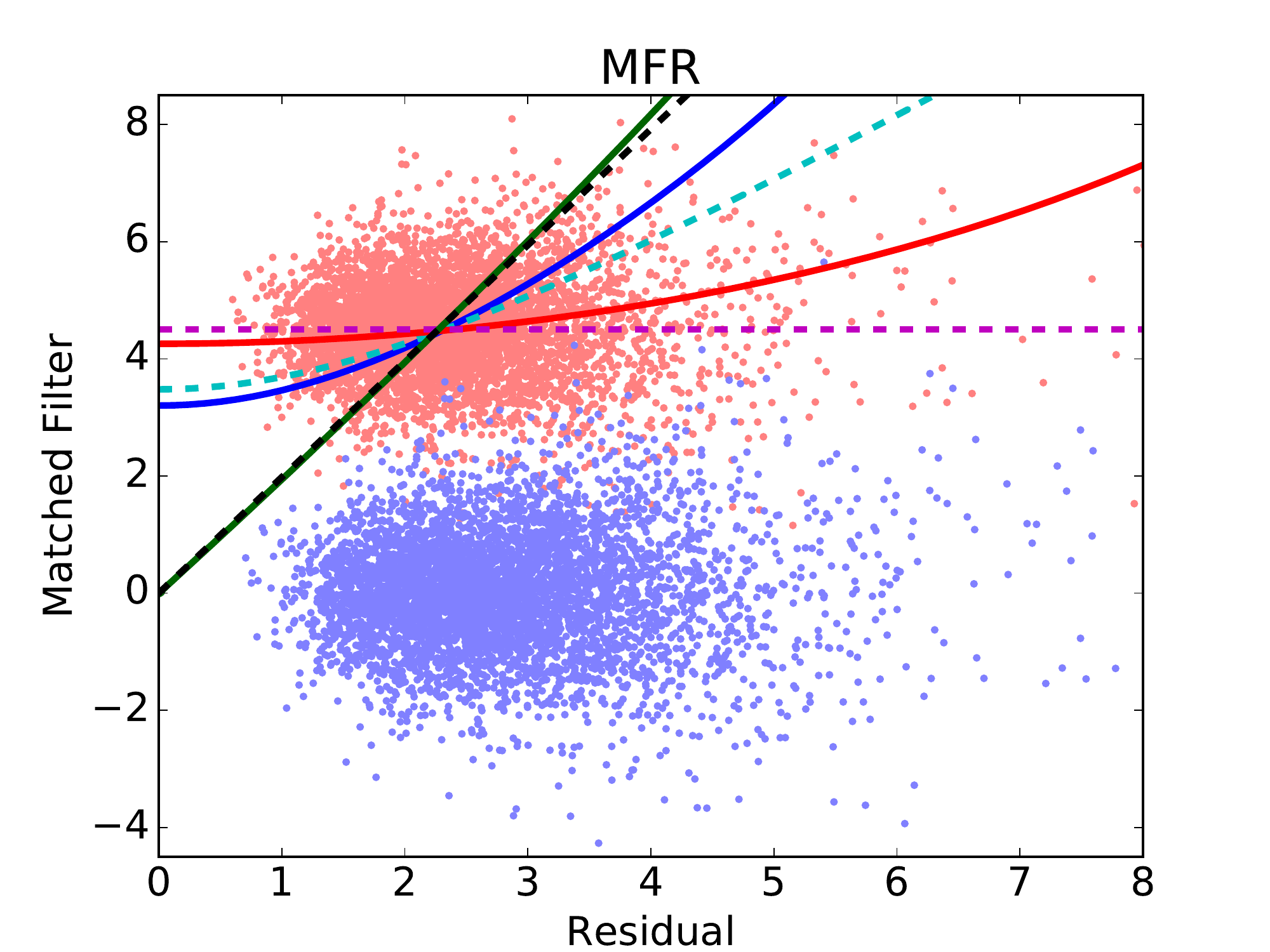}}
\centerline{\includegraphics[width=0.9\columnwidth]{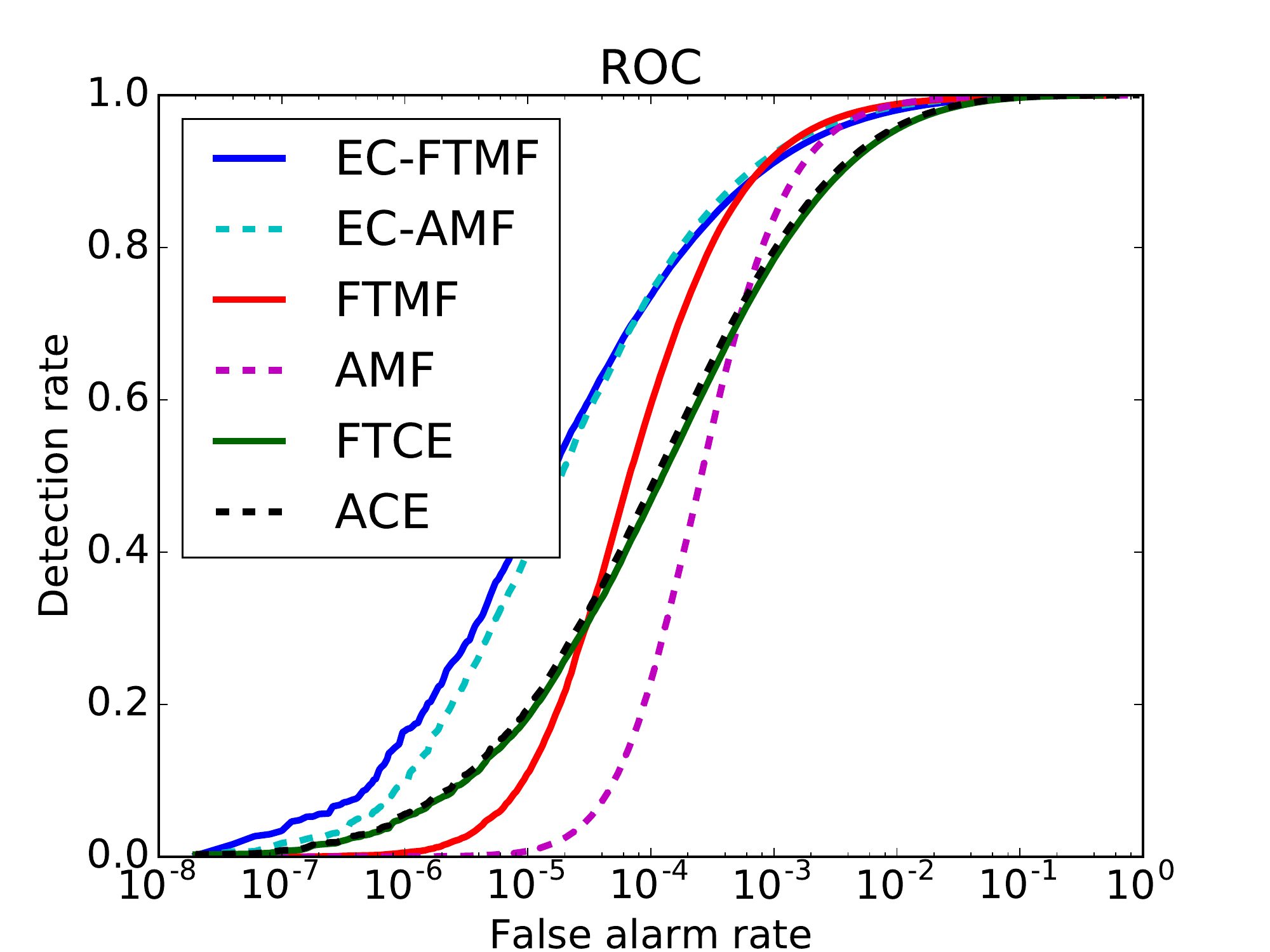}}
\caption{\label{fig:xtreme-add-2}Here, $\nu=10$, $d=10$, $T=30$, and
  $\alpha=0.15$.  The target is strong ($T\gg 1$) and small
  ($\alpha\ll 1$), so the effect due to occlusion is limited, but
  still discernible. Here, the additive-model detectors do almost (but
  not quite) as well as the corresponding replacement-model
  detectors.}
\end{figure}

The background and target pixels are $d$-dimensional vectors $\x$, but
are presented in a two-dimensional matched-filter-residual (MFR)
plot\cite{Foy09} in which the matched-filter magnitude MF is plotted
on the $y$-axis and the residual R is on the $x$-axis. 
In this zero-mean unit-covariance case:
\begin{align}
  \mbox{MF} =& \t^\T \x/T \\
  \mbox{R} = & \sqrt{\x^\T\x - (\mbox{MF})^2}.
\end{align}
\fig{mfrroc} illustrates these pixels as points in a scatter-plot.  The
reason for choosing this representation is that all of the detectors
we consider here have contours that can be plotted in this
two-dimensional space.
\fig{mfrroc} also compares the performance of various detectors on this
simulated data, and for these parameters, we see that the new EC-FTMF 
detector does well.  The original FTMF is confounded because it incorrectly
assumes the background is Gaussian; the ACE, AMF, and EC-AMF detectors are
confounded because they incorrectly assume the additive target model.

In the regime of very large $T$ and very small $\alpha$, the
replacement model is ``nearly'' additive.  In \fig{xtreme-add-2}, we
observe that the replacement-model and additive-model variants of the
same detectors are similar, although EC-FTMF is still discernibly
better than EC-AMF, and FTMF is substantially better than AMF.
Interestingly, FTCE is no better than ACE.

We have observed (results not shown here) that larger $T$ and smaller
$\alpha$ lead to a regime in which replacement-model and
additive-model variants are virtually identical.


\section{Discussion}

In introducing the EC-FTMF detector, and showing that the GLRT
solution can be expressed in closed form, we obtain a target detection
algorithm that is both convenient and adaptive to a range of
conditions. 
%
In practice, using EC-FTMF (just as in using EC-AMF or other EC-based
algorithms), one must estimate the multivariate $t$-distribution parameter $\nu$
that describes the fatness of the tails.  To keep things simple, our
simulations employed the same $\nu$ in the algorithm that was used for
the simulation. But estimation of the single scalar parameter $\nu$
from a large dataset is not that difficult; one simple approach
employs higher moments of the whitened data\cite{Theiler09ec-note}.

Finally, we remark that the GLRT -- although widely used, and very
often with good results -- is not the only or necessarily the optimal
solution to the composite hypothesis testing problem.  One may prefer
Bayesian approaches\cite{Lehmann05} or the recently introduced clairvoyant
fusion\cite{Schaum10,Theiler12spie}.

\remark{although ACE has magical properties (even though derived for additive model using
background that is Gaussian w/ unknown
variance, it is good for non-Gaussian background and for replacement model)}

\bibliographystyle{IEEEtran}
\bibliography{ecftmf}

\end{document}